\documentclass[11pt,a4paper]{article}
\usepackage[hyperref]{acl2020}
\usepackage{times}
\usepackage{latexsym}

\usepackage{microtype}
\usepackage{xspace}
\usepackage{graphicx}
\usepackage{multirow}
\usepackage{latexsym}
\usepackage{enumitem}
\usepackage{mathtools}
\usepackage{makecell}
\usepackage{amssymb}
\usepackage{amsmath, nccmath}
\usepackage{algorithm}
\usepackage[noend]{algpseudocode}
\usepackage[T1]{fontenc}

\usepackage{booktabs}
\usepackage{adjustbox}
\usepackage{csquotes}
\usepackage{url}

\usepackage{todonotes}
\makeatletter
\newcommand*\iftodonotes{\if@todonotes@disabled\expandafter\@secondoftwo\else\expandafter\@firstoftwo\fi}  
\makeatother

\everypar{\looseness=-1}

\usepackage{cleveref}
\crefname{section}{\S}{\S\S}
\Crefname{section}{\S}{\S\S}
\crefname{table}{Table}{}
\crefname{figure}{Figure}{}
\crefname{algorithm}{Algorithm}{}
\crefname{algorithm}{Algorithm}{}
\crefname{line}{Line}{}
\crefname{appendix}{App.}{}
\crefformat{section}{\S#2#1#3}
\crefname{thm}{Theorem}{}
\crefname{prop}{Proposition}{}
\crefname{def}{Definition}{}

\DeclareMathOperator{\argmax}{argmax}

\newcommand{\cell}[1]{\textsc{#1}}  
\newcommand{\cut}[1]{}
\newcommand{\word}[1]{\textit{#1}}
\newcommand{\defn}[1]{\textbf{#1}}

\newcommand{\bench}{\textsc{bench}\xspace}
\newcommand{\gold}{\textsc{gold}\xspace}
\newcommand{\superbad}{\textsc{sup}\xspace}
\newcommand{\bigo}{\mathcal{O}}

\newcommand{\Fgrid}{F_{\mathrm{grid}}}
\newcommand{\Fpar}{F_{\mathrm{par}}}
\newcommand{\Fcell}{F_{\mathrm{cell}}}
\newcommand{\score}{\mathrm{score}}

\newcommand{\decel}{\mathrm{decel}}

\aclfinalcopy 

\title{The Paradigm Discovery Problem}

\usepackage{tipa}

\newcommand{\nyu}{\normalfont \text{\textipa{D}}}
\newcommand{\ethz}{\text{\normalfont \textipa{Q}}}
\newcommand{\ohio}{\normalfont \text{\textipa{@}}}
\newcommand{\jhu}{\normalfont \text{\textipa{Z}}}
\newcommand{\ucambridge}{\normalfont \text{\textipa{6}}}

\author{Alexander Erdmann$^{\ohio,\nyu}$~\;~Micha Elsner$^{\ohio}$~\;~Shijie Wu$^{\jhu}$ \\ \textbf{Ryan Cotterell}$^{\ethz,\ucambridge}$~\;~\textbf{Nizar Habash}$^{\nyu}$ \\
$^{\ohio}$Ohio State University~\;~$^{\nyu}$New York University Abu Dhabi\\
$^{\jhu}$Johns Hopkins University~\;~$^{\ucambridge}$University of Cambridge~\;~$^{\ethz}$ETH Z{\"u}rich\\
\texttt{\{ae1541,nizar.habash\}@nyu.edu}~\;~\texttt{elsner.14@osu.edu}\\
\texttt{shijie.wu@jhu.edu}~\;~\texttt{ryan.cotterell@inf.ethz.ch}}

\date{}

\begin{document}
\maketitle

\begin{abstract}
This work treats the paradigm discovery problem (PDP)---the task of learning an inflectional morphological system from unannotated sentences.
We formalize the PDP and develop evaluation 
metrics for judging systems.
Using currently available resources, we
construct datasets for the task. We also devise a heuristic benchmark for the PDP and report empirical results on five diverse languages.
Our benchmark system first makes use of word embeddings and string similarity to cluster forms by cell and by paradigm.
Then, we bootstrap a neural transducer on top of the clustered data to predict words to realize the empty paradigm slots. An error analysis of our system suggests clustering by cell across different inflection classes is the most pressing challenge for future work. 
Our \href{https://github.com/alexerdmann/ParadigmDiscovery}{code and data} are publicly available.
\end{abstract}

\section{Introduction}\label{intro}

In childhood, we induce our native language's morphological system from unannotated input.
For instance, we learn that \word{ring} and \word{rang} belong to the same inflectional paradigm. 
We also learn that \word{rings} and \word{bangs} belong to the same cell, i.e., they realize the same morphosyntactic properties \cell{3.sg.pres}, but in different paradigms.
Acquiring such paradigmatic knowledge enables us to produce unseen inflectional variants of new vocabulary items, i.e. to complete morphological paradigms. Much work has addressed this task, which \newcite{ackerman2009parts} call the paradigm cell filling problem (PCFP),\footnote{In the NLP literature, this task is called morphological reinflection or morphological inflection generation \cite{cotterell-sigmorphon2016}; this is only a difference in nomenclature.} 
but few have discussed inducing paradigmatic knowledge from scratch, 
which we call the \textbf{paradigm discovery problem} (PDP).\footnote{\newcite{elsner2019modeling} call the task the paradigm cell discovery problem; we drop \emph{cell} to distinguish our task from one of its subtasks which \newcite{boye2019realistic} call the paradigm cell finding problem (see \cref{subtasks}).} 

As an unsupervised task, the PDP poses challenges for modeling and evaluation and has yet to be attempted in its full form \cite{elsner2019modeling}. 
However, we contend there is much to be gained from formalizing and studying the PDP. 
There are insights for cognitive modeling to be won \cite{pinker2001four,goldwater2007nonparametric} and intuitions on combating sparse data for language generation \cite{King18} to be accrued. Unsupervised language processing also has natural applications in the documentation of endangered languages \cite{zamaraeva2019handling}
where a lot of annotated data is never likely to exist.
Our formalization of the PDP offers a starting point for future work on unsupervised morphological paradigm completion. 

Our paper presents a concrete formalization of the PDP.
Then, as a baseline for future work, we introduce a heuristic benchmark system.
Our benchmark system takes an unannotated text corpus and a lexicon of words from the corpus to be analyzed.
It first clusters the lexicon by cell and then by paradigm making use of distributional semantics and string similarity. 
Finally, it uses this clustering as silver-standard supervision to bootstrap a neural transducer \cite{vaswani2017attention} that generates the desired target inflections.
That is, the model posits forms to realize unoccupied cell slots in each proposed paradigm.
Even though our benchmark system models only one part of speech (POS) at a time, our framework extends to the full PDP to support future, more intricate systems.
We propose two separate metrics to evaluate both the clustering of attested forms into paradigms and cells and the prediction of unseen inflected forms.
Our metrics handle non-canonical morphological behavior discussed in theoretical literature \cite{corbett2005canonical} and extend to the full PDP.

For three of the five languages we consider, our benchmark system predicts unattested inflections of lexicon forms with accuracy within 20\% of a fully supervised system.
However, our analysis suggests clustering forms into cells consistently across paradigms is still a very pressing challenge.


\section{Previous Work in Morphology}


This section couches our work on the PDP in terms of previous trends in morphological modeling.

\subsection{Unsupervised Morphology}
\label{unsupworks}

Much work on unsupervised morphological modeling focuses on segmentation  \cite{gaussier-1999-unsupervised,goldsmith2001unsupervised,creutz2005unsupervised,narasimhan2015unsupervised,bergmanis2017segmentation,xuu}.
While morphological segmenters can distinguish real from spurious affixes (e.g., \word{bring} $\neq$ \word{br} + \word{ing}) with high accuracy, they do not attempt to solve the PDP. They do, however, reveal which forms take the same affixes (e.g., \word{walked}, \word{talked}), not which forms occupy the same cell (e.g., \word{walked}, \word{brought}). Indeed, they explicitly struggle with irregular morphology. Segmenters also cannot easily model non-concatenative phenomena like ablaut, vowel harmony and templatic processes.

Two works have proposed tasks which can be considered alternative formulations of the PDP, using either minimal or indirect supervision to bootstrap their models.
We discuss each in turn.
First, \newcite{dreyer2011discovering} use a 
generative model to cluster forms into paradigms and cells with a Bayesian non-parametric mixture of weighted finite-state transducers.
They present a PDP framework which, in principle, could be fully unsupervised, but their model requires a small seed of labeled data to get key information like the number of cells distinguished, making it less relevant cognitively.
In contrast, our task is not directly supervised and focuses on distributional context.
Second, contemporaneous to our work, \newcite{sharedtask_2_2020} propose a similar framework for SIGMORPHON 2020's \href{https://sigmorphon.github.io/sharedtasks/2020/task2}{shared task} on unsupervised morphological paradigm completion.
Given only a small corpus and lexicon of verbal lemmata, participating systems must propose full paradigms for each lemma.
By contrast, our framework does not reveal how many paradigms should be generated, nor do we privilege a specific form as the lemma, but we do use a larger lexicon of exclusively verbal or nominal forms.
Their proposed baseline uses distributional context for POS tagging and features, but does not train embeddings as the corpus is small.

\begin{table*}[]
\centering
\resizebox{0.65\textwidth}{!}{%
\begin{tabular}{cccccc}
\multirow{3}{*}{\textbf{Corpus}} & \multicolumn{5}{c}{\textit{The cat \underline{watched} me \underline{watching} it .}} \\
 & \multicolumn{5}{c}{\textit{I \underline{followed} the show but she had n't \underline{seen} it .}} \\
 & \multicolumn{5}{c}{\textit{Let 's \underline{see} who \underline{follows} your logic .}} \\
 &  &  &  &  &  \\
\textbf{Lexicon} & \multicolumn{5}{c}{\word{watching}, \word{seen}, \word{follows}, \word{watched}, \word{followed}, \word{see}} \\
 &  &  &  &  &  \\
\textbf{Gold Grid} & cell 1 & cell 2 & cell 3 & cell 4 & cell 5 \\ \cline{2-6} 
paradigm 1 & \guillemotleft{}\word{watch}\guillemotright{} & \guillemotleft{}\word{watches}\guillemotright{} & \word{watching} & \word{watched} & \word{watched} \\ \cline{2-6} 
paradigm 2 & \guillemotleft{}\word{follow}\guillemotright{} & \word{follows} & \guillemotleft{}\word{following}\guillemotright{} & \word{followed} & \word{followed} \\ \cline{2-6} 
paradigm 3 & \word{see} & \guillemotleft{}\word{sees}\guillemotright{} & \guillemotleft{}\word{seeing}\guillemotright{} & \guillemotleft{}\word{saw}\guillemotright{} & \word{seen} \\ \cline{2-6} 
\end{tabular}%
}
\caption{An example corpus, lexicon, and gold analyses. All lexicon entries appear in the corpus and, for our experiments, they will all share a POS, here, verb. The grid reflects all possible analyses of syncretic forms (e.g., {\it walked}, {\it followed}), even though these only occur in the corpus as \cell{pst} realizations, like {\it saw} in Cell 4, not as \cell{pst.ptcp}, like {\it seen} in Cell 5. Bracketed \guillemotleft{}forms\guillemotright{} are paradigm mates of attested forms, not attested in the lexicon.}
\label{toy}
\end{table*}


\subsection{Subtasks of Paradigm Discovery}
\label{subtasks}

A few works address subtasks of the PDP.
\newcite{erdmann2018complementary} learn paradigm membership from raw text, but do not sort paradigms into cells.
\newcite{boye2019realistic} discuss the paradigm cell finding problem, identifying the cell (but not paradigm) realized by a given form.
\newcite{lee-2015-morphological} clusters forms into cells across inflection classes.
\newcite{beniamine2018inferring} group paradigms into inflection classes, and \newcite{eskander2013automatic} induce inflection classes and lemmata from cell labels.

\subsection{The Paradigm Cell Filling Problem}
\label{supworks}

The PCFP is the task of predicting unseen inflected forms given morphologically labeled input.
PCFP models can guess a word's plural having only seen its singular, but the child must bootstrap morphological knowledge from scratch, first learning that singular--plural is a relevant distinction.
Thus, the PDP must be at least partially solved before the PCFP can be attempted.
Yet, as a supervised task, the PCFP is more easily studied, and has received much attention on its own, especially from the word-and-paradigm camp of morphological theory.

Some cognitive works suggest the PCFP cannot be too difficult for any language \cite{dale-etal-1998-introduction,ackerman2013morphological,AckermanMalouf15,blevins2017zipfian,cotterell2019complexity}.
Neural models can test and extend such proposals \cite{cotterell2018diachronic,silfverberg2018encoder}.
A related vein of work discusses how speakers inflect nonce words \cite{berko58,plunkett1999connectionist,yang2015}, e.g., is the past tense of \word{sping}, \word{spinged} or \word{spung}? There is a long 
tradition of modeling past-tense generation with neural networks \cite{rumelhart1986learning,kirov2018recurrent,Corkery19}.

On the engineering side, \newcite{durrett2013supervised} inspired much recent work, which has since benefited from large inflectional datasets \cite{kirov2018unimorph} and advances in neural sequence modeling \cite{bahdanau2014neural}.
Shared tasks have drawn extra attention
to the PCFP \cite{cotterell-sigmorphon2016, cotterell2017conll,cotterell2018conll,mccarthy2019sigmorphon}.

 \label{rw}

\section{The Paradigm Discovery Problem}\label{task}

Paradigm discovery is a natural next step in computational morphology, building on related minimally or indirectly supervised works (\cref{subtasks}) to bridge the gap between unsupervised traditions (\cref{unsupworks}) and supervised work on the PCFP (\cref{supworks}).
In the PCFP, each input form is labeled with its morphosyntactic property set, i.e., the \defn{cell} in the paradigm which it realizes, and its lexeme, i.e., the \defn{paradigm} of related forms to which it belongs.
By contrast, to solve the PDP, unlabeled input forms must be assigned cells and paradigms.
This task requires learning what syntactic and semantic factors distinguish cells, what combinations of cells can co-occur in a paradigm, and what aspects of a surface form reflect its paradigm and its cell, respectively.

\subsection{Task Setup}
\label{tasksetup}

\cref{toy} provides an overview of our PDP setup. 
The first two rows show input data: an unannotated \defn{corpus} and a \defn{lexicon} of forms attested in that corpus.
Given only these data, the task is to output a \defn{grid} such that (i) all lexicon forms and all their (potentially unseen) inflectional variants appear in the grid, (ii) all forms appearing in the same column realize the same morphosyntactic cell, and (iii) all forms appearing in the same row belong to the same paradigm.
Unattested \guillemotleft{}\word{forms}\guillemotright{} to be generated are depicted in brackets in \cref{toy}'s Gold Grid, which shows the ideal output of the system.

Our setup permits multiple forms realizing the same \defn{slot}, i.e., a specific cell in a specific paradigm, a single form realizing multiple slots, and unrealizable empty slots.
This supports overabundance \cite{thornton2010towards, thornton2011overabundance}, defectiveness \cite{sims2015inflectional}, and syncretism \cite{blevins1995syncretism, cotterell2018unsupervised}.
See \newcite{corbett2005canonical} for more on these phenomena.
Experimentally, we constrain the PDP by limiting the lexicon to forms from one POS, but our formalization is more general.


\subsection{Data for the PDP}
\label{data}

For a given language and POS, we create a corpus, lexicon, and gold grid based on a Universal Dependencies (UD) corpus \cite{nivre2016universal}. At a high level, the corpus includes raw, non-UD sentences, and UD sentences stripped of annotations.
The lexicon includes all forms occurring in the UD sentences with the specified POS (potentially including variant spellings and typographical errors).
The gold grid consists of full paradigms for every word which co-occurs in UD and the UniMorph lexicon \cite{kirov2018unimorph} with a matching lemma--cell analysis; this is similar to the corpus created by \newcite{vylomova-etal-2019-contextualization}.
As a system does not know which lexicon forms will be evaluated in the gold grid, it must model the entire lexicon, which should contain a realistic distribution over rare words and inflection classes having been directly extracted from distributional data \cite{Bybee03,Lignos18}.\looseness=-1



\begin{table*}
\centering
\resizebox{0.63\textwidth}{!}{%
\begin{tabular}{ccccc}
\textbf{Predictions} & cell 1 & cell 2 & cell 3 & cell 4 \\ \cline{2-5} 
paradigm 1 & \word{watched} & \word{watching} & \guillemotleft{}\word{watches}\guillemotright{} & \guillemotleft{}\word{watch}\guillemotright{} \\ \cline{2-5} 
paradigm 2 & \word{followed} & \guillemotleft{}\word{following}\guillemotright{} & \word{follows} & \guillemotleft{}\word{follow}\guillemotright{} \\ \cline{2-5} 
paradigm 3 & \guillemotleft{}\word{seed}\guillemotright{} & \guillemotleft{}\word{seeing}\guillemotright{} & \guillemotleft{}\word{sees}\guillemotright{} & \word{see} \\ \cline{2-5} 
paradigm 4 & \guillemotleft{}\word{seened}\guillemotright{} & \guillemotleft{}\word{seening}\guillemotright{} & \guillemotleft{}\word{seens}\guillemotright{} & \word{seen} \\ \cline{2-5} 
\end{tabular}%
}
\caption{Toy predictions made from the corpus and lexicon in \cref{toy}, to be evaluated against the toy gold grid. Again, bracketed \guillemotleft{}forms\guillemotright{} are those not occurring in the lexicon.}
\label{toy_pred}
\end{table*}

To ensure the gold grid is reasonably clean, we take all word--lemma--feature tuples from the UD portion of the corpus matching the specified POS and convert the features to a morphosyntactic cell identifier compatible with UniMorph representation as in \newcite{mccarthy-etal-2018-marrying}.%
\footnote{Aligning UniMorph and UD requires removing diacritics in (Latin and Arabic) UniMorph corpora to match UD. 
This can obscure some morphosyntactic distinctions but is more consistent with natural orthography in distributional data.
The use of orthographic data for morphological tasks is problematic, but standard in the field, due to scarcity of phonologically transcribed data \cite{ackerman19databases}.}
Then we check which word--lemma--cell tuples also occur in UniMorph.
For each unique lemma in this intersection, the full paradigm is added as a row to the gold grid.
To filter typos and annotation discrepancies, we identify any overabundant slots, i.e., slots realized by multiple forms, and remove all but the most frequently attested realization in UD.
While some languages permit overabundance \cite{thornton2010towards}, it often indicates typographical or annotation errors in UD and UniMorph \cite{gorman19,ackerman19databases}.
Unlike the gold grid, the lexicon retains overabundant realizations, requiring systems to handle such phenomena.


For each language, the raw sentences used to augment the corpus add over 1 million additional words.
For German and Russian, we sample sentences from OpenSubtitles \cite{lison2016opensubtitles2016}, for Latin, the Latin Library \cite{johnson2016cltk}, and for English and Arabic, Gigaword \cite{LDC:Gigaword-5,parker2011english}.
Supplementary sentences are preprocessed via Moses \cite{koehn-etal-2007-moses} to split punctuation, and, for supported languages, clitics.
\cref{stats} shows corpus and lexicon sizes.

\subsection{Metrics} \label{metrics}
A system attemping the PDP is expected to output a morphologically organized grid in which rows and columns are arbitrarily ordered, but ideally, each row corresponds to a gold paradigm and each column to a gold cell.
Aligning rows to paradigms and columns to cells is non-trivial, making it difficult to simply compute accuracy over gold grid slots.
Furthermore, cluster-based metrics \cite{rosenberg2007v} are difficult to apply as forms can appear in multiple columns or rows.
Thus, we propose novel metrics that are \emph{lexical}, based on analogical relationships between forms.
We propose a set of PDP metrics, to measure how well organized lexicon forms are in the grid, and a set of PCFP metrics, to measure how well the system anticipates unattested inflectional variants.
All metrics support non-canonical phenomena such as defective paradigms and overabundant slots.

\subsubsection{PDP Metrics}

A form $f$'s \defn{paradigm mates} are all those forms that co-occur in at least one paradigm with $f$.
$f$'s \defn{paradigm F-score} is the harmonic mean of precision and recall of how well we predicted its paradigm mates when viewed
as an information retrieval problem \cite{manning2008introduction}. 
We macro-average all forms' paradigm F-scores to compute $\Fpar$. 
Qualitatively, $\Fpar$ tells us how well we cluster words that belong to the \emph{same paradigm}. 
A form $f$'s \defn{cell mates} are all those forms that co-occur in at least one cell with $f$. 
$f$'s \defn{cell F-score} is the harmonic mean of precision and recall of how well we predicted its cell mates. 
As before, we macro-average all forms' cell F-scores to compute $\Fcell$. Qualitatively, $\Fcell$ tells us how well we cluster words that belong to the \emph{same cell}. 
Finally, we propose the $\Fgrid$ metric as the harmonic mean of $\Fpar$ and $\Fcell$.
$\Fgrid$ is a single number that reflects a system's ability to cluster forms into \emph{both} paradigms and cells.
Because we designate separate PCFP metrics to evaluate gold grid forms not in the lexicon, we restrict $f$'s mates to only include forms that occur in the lexicon.

Consider the proposed grid in \cref{toy_pred}.
There are 6 lexicon forms in the gold grid.
Starting with \word{watched}, we correctly propose its only attested paradigm mate, \word{watching}.
Thus, \word{watched}'s paradigm F-score is 100\%.
For \word{see}, we propose no attested paradigm mates, but we should have proposed \word{seen}.
0 correct out of 1 true paradigm mate from 0 predictions results in an F-score of 0\% for \word{seen}. 
We continue like this for all 6 attested forms in the gold grid and average their scores to get $\Fpar$.
As for $\Fcell$, we correctly predict that \word{watched}'s only cell mate is \word{followed}, yielding an F-score of 100\%.
However, we incorrectly predict that \word{see} has a cell mate, \word{seen}, yielding an F-score of 0\%; we average each word's F-score to get $\Fcell$; the harmonic mean of $\Fpar$ and $\Fcell$ gives us $\Fgrid$.

While $\Fgrid$ handles syncretism, overabundance, defectiveness and mismatched grid dimensions, it is exploitable by focusing exclusively on the best attested cells realized by the most unique forms, since attested cells tend to exhibit a Zipfian distribution \cite{blevins2017zipfian,Lignos18}.
Exploiting $\Fgrid$ in this manner propagates errors when bootstrapping to predict unattested forms and, thus, will be punished by PCFP metrics.

\begin{table}
\centering
\begin{tabular}{rccc} \toprule
\multicolumn{1}{c}{} & \textbf{Lexicon}  & \textbf{Corpus} & \textbf{UD} \\ \midrule
\multicolumn{1}{c}{} & \textbf{Types}  & \textbf{Token} & \textbf{Tokens} \\
\cline{2-4}
\multicolumn{1}{r}{Arabic} & 8,732 & 1,050,336 & 223,881 \\ 
\multicolumn{1}{r}{German} & 19,481 & 1,270,650 & 263,804 \\
\multicolumn{1}{r}{English} & 3,330 & 1,212,986 & 204,608 \\ 
\multicolumn{1}{r}{Latin} & 6,903 & 1,218,377 & 171,928 \\
\multicolumn{1}{r}{Russian} & 36,321 & 1,885,302 & 871,548 \\ \bottomrule
\end{tabular}%
\caption{Statistics regarding the input corpus and lexicon. UD tokens refers to tokens in the corpus originally extracted from UD sentences.}
\label{stats}
\end{table}
\subsubsection{PCFP Metrics}

We cannot evaluate the PCFP as in supervised settings \cite{cotterell-sigmorphon2016} because proposed cells and paradigms cannot be trivially aligned to gold cells and paradigms.
Instead, we create a test set by sampling 2,000 four-way analogies from the gold grid.
The first and second forms must share a row, as must the third and fourth; the first three forms must be attested and the fourth unattested, e.g.,
\word{watched} : \word{watching} :: \word{seen} : \guillemotleft{}\word{seeing}\guillemotright{}.

From this test set and a proposed grid, we compute a strict Analogy (An) accuracy metric and a lenient Lexicon Expansion (LE) accuracy metric.
Analogy counts instances as correct if \emph{all} analogy directions hold in the proposed grid (i.e., \word{watched}, \word{watching} and \word{seen}, \guillemotleft{}\word{seeing}\guillemotright{} share rows and \word{watched}, \word{seen} and \word{watching}, \guillemotleft{}\word{seeing}\guillemotright{} share columns).
Lexicon Expansion counts instances as correct if the unattested fourth form appears anywhere in the grid. That is, Lexicon Expansion asks, for each gold form, if it was predicted in \emph{any} slot in \emph{any} paradigm. 

Like the PDP metrics, our PCFP metrics support syncretism, overabundance, defectiveness, etc.
One can, however, exploit them by proposing a gratuitous number of cells, paradigms, and syncretisms, increasing the likelihood of completing analogies by chance, though this will reduce $\Fgrid$. As both PDP and PCFP metrics can be exploited independently but not jointly, we argue that both types of metrics should be considered when evaluating an unsupervised system.

\section{Building a Benchmark}\label{model}

This section presents a benchmark system for proposing a morphologically organized grid given a corpus and lexicon.
First, we cluster lexicon forms into cells.
Then we cluster forms into paradigms given their fixed cell membership.
To maintain tractability, clustering assumes a one-to-one mapping of forms to slots.
Following cell and paradigm clustering, we predict forms to realize empty slots given one of the lexicon forms assigned to a cell in the same paradigm.
This allows forms to appear in multiple slots, but does not support overabundance, defectiveness, or multi-word inflections.

\subsection{Clustering into Cells}
\label{cell_clustering}

We use a heuristic method to determine the number of cells and what lexicon forms to assign to each.
Inspired by work on inductive biases in word embeddings \cite{pennington2014glove,trask2015modeling,goldberg2016primer,avraham2017interplay,tu2017learning}, we train morphosyntactically biased embeddings on the corpus and use them to $k$-means cluster lexicon forms into cells.
Following \newcite{erdmann2018addressing}, we emphasize morphosyntactically salient dimensions in embedding space by manipulating hyperparameters in FastText \cite{bojanowski2016enriching}.
Specifically, to encourage grouping of morphologically related words, FastText computes a word's embedding as the sum of its subword embeddings for all subword sequences between 3 and 6 characters long \cite{Schuetze93wordspace}.
We shorten this range to 2 to 4 to bias the grouping toward shared affixes rather than (usually longer) shared stems.
This helps recognize that the same affix is likely to realize the same cell, e.g., \word{watch +ed} and \word{follow +ed}.
We limit the context window size to 1; small windows encourage a morphosyntactic bias in embeddings \cite{erk2016alligator}.

We determine the number of cells to cluster lexicon forms into, $k$, via the \defn{elbow method}, which progressively considers adding clusters until the reduction in dispersion levels off \cite{kodinariya2013review,bholowalia2014ebk}.\footnote{Clustering dispersion is the squared distance of a point from its cluster's centroid, summed over all points clustered.}
Since \newcite{tibshirani2001estimating}'s popular formalism of the method does not converge on our data, we implement a simpler technique that works in our case.
We incrementally increase $k$, each time recording clustering dispersion, $d_{k}$ (for consistency, we average $d_{k}$ over 25 iterations).
Starting at $k=2$, we calculate dispersion deceleration as the difference between the current and previous dispersions:
\begin{equation} \label{dd}
\decel(k) = d_{k-1}-2(d_{k})+d_{k+1}
\end{equation}
Once $\decel(k)$ decreases below $\sqrt{\decel(2)}$, we take the $k^{\text{th}}$ clustering: the $(k+1)^{\text{th}}$ cluster did not explain enough variation in the embedding space to justify an additional morphosyntactic distinction.

\subsection{Clustering into Paradigms}

Given a clustering of lexicon forms into $k$ cells, denoted as $C_1, \ldots, C_K$, we heuristically cluster each form $f$ into a paradigm, $\pi$, as a function of $f$'s cell, $c$.
For tractability, we assume paradigms are pairwise disjoint and no paradigm contains multiple forms from the same cell, greedily building paradigms cell by cell. 
To gauge the quality of a candidate paradigm, we first identify its \defn{base} and \defn{exponents}.
Following \newcite{beniamine2018inferring}, we define $\pi$'s base, $b_\pi$, as the longest common subsequence shared by all forms in $\pi$.\footnote{The fact that we use a sub\emph{sequence}, instead of a sub\emph{string}, means that we can handle non-concatenative morphology.}%
\footnote{We note that the longest common subsequence may be found with a polynomial-time dynamic program; however, there will not exist an algorithm whose runtime is polynomial in the \emph{number of strings} unless $\textsf{P} = \textsf{NP}$ \cite{maier1978complexity}.}
For each form $f$ in $\pi$, we define the exponent $x_f$ as the subsequences of $f$ that remain after removing $b_\pi$, i.e., $x_f$ is a tuple of affixes.
For example, if $\pi$ contains words \word{wxyxz} and \word{axx}, $b_\pi$ is \word{xx} and the exponents are (\word{<w}, \word{y}, \word{z>}) and (\word{<a}), respectively.\footnote{While we use word start (<) and end (>) tokens to distinguish exponents, they do not count toward the number of characters in \cref{par_score}.}
Inspired by unsupervised maximum matching in greedy tokenization \cite{guo1997critical,erdmann2019little}, we define the following paradigm score function:
\begin{equation} \label{par_score}
\score(\pi) =  \sum_{\langle c, f \rangle \in\pi}\Big(|b_\pi| - |x_f| \Big)
\end{equation}
which scores a candidate paradigm according to the number of base characters minus the number of exponent characters (it can be negative).
\cref{pc} then details our heuristic clustering approach from start to finish, as we greedily select one or zero forms from each cell to add (via the list concatenation operator $\circ$) to each paradigm such that the paradigm's score is maximized.%
\footnote{\cref{pc} has complexity $\bigo(|L|^2)$ where $|L|$ is lexicon size. In practice, to make \cref{pc} tractable, we limit the candidates for $f_j'$ (line 8) to the $n=250$ forms from cell $j$ nearest to $f_i$ in pre-trained embedding space (trained via FastText with default parameters). This achieves a complexity upper bounded by $\bigo(|L|nk)$.}
\begin{algorithm}
	\caption{Paradigm Clustering Algorithm}
	\label{pc}
	\begin{algorithmic}[1]
	    \State \textbf{input} ${C_1,\ldots,C_k}$
	    \State ${\boldsymbol \pi} \gets [\,]$
		\For {$C_i \in \{C_1,\ldots,C_k\}$}
			\For {$f_i \in C_i$}
				\State $\pi \gets [\langle i, f_i \rangle]$ 
				\State $s \gets \score(\pi)$
				\For {$C_j \in \{C_{i+1},\ldots,C_k\}$}
				    \State $f_j\gets \underset{f_j' \in C_j}{\argmax}\,\,\score(\pi \circ [\langle j, f_j' \rangle])$
				    \State $s_{f_j}\gets \score(\pi \circ [\langle j, f_j \rangle])$
				    \If {$s_{f_j} > s$}
				        \State $\pi \gets \pi \circ [\langle j, f_j\rangle]$ 
				        \State $s \gets s_{f_j}$
			            \State  $C_j.\mathrm{remove}(f_j)$
				    \EndIf
				\EndFor
		        \State ${\boldsymbol \pi} \gets {\boldsymbol \pi} \circ [\pi]$
		      \EndFor
		   \EndFor
	\State \textbf{return} ${\boldsymbol \pi}$
	\end{algorithmic} 
\end{algorithm}

After performing a first pass of paradigm clustering with \cref{pc}, we estimate an unsmoothed probability distribution $p(x \mid c)$ as follows: we take the number of times each exponent (tuple of affixes) realizes a cell in the output of \cref{pc} and divide by the number of occurrences of that cell.
We use this distribution $p(x \mid c)$ to construct an exponent penalty:
\begin{align} \label{penalty}
\omega(&x_f,c)  \\
&=\begin{cases}
    0& \hspace{-.2cm}\textbf{if } \underset{x}{\argmax}\,\,p(x \mid c) = x_f\\
    2 - \frac{p(x_f \mid c)}{\max_{x}p(x \mid c)}  & \hspace{-.2cm}\textbf{otherwise} \nonumber
\end{cases}
\end{align}
Intuitively, if an exponent is the most likely exponent in the cell to which it belongs, the penalty weight is zero and its characters are not subtracted from the score.
Otherwise, the weight is in the interval $[1, 2]$ such that each exponent character is penalized at least as harshly but no more than twice as harshly than in the first pass, according to the exponent's likelihood.
We use this exponent penalty weight to define a penalized score function:
\begin{equation} \label{newscore}
\score_\omega(\pi) =  \sum_{\langle c, f \rangle \in\pi}\Big(|b_\pi| - |x_f|\, \omega(x_f,c) \Big)
\end{equation}

We then re-run \cref{pc}, swapping out $\score(\cdot)$ for $\score_\omega(\cdot)$, to re-cluster forms into paradigms. 
Empirically, we find that harsher exponent penalties---i.e., forcing weights to be greater than 1 for suboptimal exponents---lead to higher paradigm precision in this second pass. For an example, consider candidate paradigm \mbox{[<<>>, {\it watched}, <<>>, <<>>, <<>>]}.
If we add nothing, each character of \textit{watched} can be analyzed as part of the base, yielding a score of 7.
What if we attempt to add \textit{watching}---pre-determined to belong to column 5 during cell clustering?
Candidate paradigm \mbox{[<<>>, {\it watched}, <<>>, <<>>, {\it watching}]} increases the number of base characters to 10 ({\it watch} shared by 2 words), but yields a score of 5 after subtracting the characters from both exponents, ({\it ed}>) and ({\it ing}>).
Hence, we do not get this paradigm right on our first pass, as $5 < 7$.
Yet, after the first pass, should ({\it ed}>) and ({\it ing}>) be the most frequent exponents in the second and fifth cells, the second pass will be different. 
Candidate paradigm \mbox{[<<>>, {\it watched}, <<>>, <<>>, {\it watching}]} is not penalized for either exponent, yielding a score of 10, thereby allowing \word{watching} to be added to the paradigm.

\subsection{Reinflection}
\label{inflection}
We now use the output of the clustering by cell and paradigm to bootstrap the PCFP. We
use a Transformer \cite{vaswani2017attention} to predict the forms that realize empty slots. Transformer-based
neural transducers constitute the state of the art for the PCFP. 
\footnote{We use the following hyperparameters: $N=4$, $d_\textrm{{\it model}}=128$, $d_\textrm{{\it ff}}=512$. 
Remaining hyperparameters retain their default values as specified in \newcite{vaswani2017attention}.
Our models are trained for 100 epochs in batches of 64.
We stop early after 20 epochs without improvement on the development set.}
In \newcite{cotterell-etal-2016-morphological}'s terms, we reinflect the target from one of the non-empty source cells in the same paradigm. We select the source from which we can most reliably reinflect the target.
We quantify this reliability by calculating the accuracy with which each target cell's realizations were predicted from each source cell's realizations in our development set.
For each target cell, we rank our preferred source cells according to accuracy.

To generate train and development sets, we create instances for every possible pair of realizations occurring in the same paradigm (90\% train, 10\% development).
We pass these instances into the Transformer, flattening cells and characters into a single sequence.
Neural models for reinflection often perform poorly when the training data are noisy. 
We mitigate this via the harsh exponent penalty weights (\cref{penalty}) which encourage high paradigm precision during clustering.

\section{Results and Discussion}

\begin{table}
\centering
\resizebox{0.99\columnwidth}{!}{%
\setlength{\tabcolsep}{3pt}
\begin{tabular}{rcc|ccc|cc}
\toprule
    \multicolumn{1}{c}{\multirow{2}{*}{
    }} & \multicolumn{2}{c}{\textbf{}} & \multicolumn{3}{c}{\textbf{PDP}} & \multicolumn{2}{c}{\textbf{PCFP}} \\
\multicolumn{1}{c}{} & Cells & Paradigms & $\Fcell$ & $\Fpar$ & $\Fgrid$ & An & LE \\  [1mm]

\multicolumn{8}{c}{\textbf{Arabic nouns} -- 8,732 forms} \\
\hline
\multicolumn{1}{r}{\superbad} & 27 & 4,283 &  &  &  & 85.9 & 87.0 \\ 
\cline{1-1}
\multicolumn{1}{r}{\bench} & 12.8 & 5,279.3	& 39.9	&	48.5	&	43.7	&	16.8	&	49.5 \\
\multicolumn{1}{r}{\gold $k$} & 27 &	4,930.3 &	25.9 &	46.4 &	33.1 &	16.1 &	57.2 \\ [1mm]

\multicolumn{8}{c}{\textbf{German nouns} -- 19,481 forms} \\
\hline
\multicolumn{1}{r}{\superbad} & 8 & 17,018 &  &  &  & 72.2 & 74.9 \\ 
\cline{1-1}
\multicolumn{1}{r}{\bench} & 7.3 & 17,073.3 &	35.2 &	59.4 &	43.3 &	14.2 &	56.7 \\
\multicolumn{1}{r}{\gold $k$} & 8 &	16,836.0 &	29.4 &	66.6 &	40.8 &	14.8 &	60.4 \\ [1mm]

\multicolumn{8}{c}{\textbf{English verbs} -- 3,330 forms} \\
\hline
\multicolumn{1}{r}{\superbad} & 5 & 1,801 &  &  &  & 80.4 & 80.7 \\ 
\cline{1-1}
\multicolumn{1}{r}{\bench} & 7.5 &	1,949.5 &	64.0 &	80.1 &	71.1 &	52.0 &	67.5 \\
\multicolumn{1}{r}{\gold $k$} & 5 &	1,977.3 &	79.6 &	82.1 &	80.8 &	54.7 &	69.4 \\ [1mm]

\multicolumn{8}{c}{\textbf{Latin nouns} -- 6,903 forms} \\
\hline
\multicolumn{1}{r}{\superbad} & 12 & 3,013 &  &  &  & 80.0 & 88.0  \\ 
\cline{1-1}
\multicolumn{1}{r}{\bench} & 13.0 &	3,746.5 &	38.8 &	73.2 &	50.6 &	17.2 &	72.9 \\
\multicolumn{1}{r}{\gold $k$} & 12 &	3,749.0 &	39.9 &	71.6 &	51.3 &	17.5 &	72.6 \\ [1mm]

\multicolumn{8}{c}{\textbf{Russian nouns} -- 36,321 forms} \\
\hline
\multicolumn{1}{r}{\superbad} & 14 & 14,502 &  &  &  & 94.7 & 96.8 \\ 
\cline{1-1}
\multicolumn{1}{r}{\bench} & 16.5 &	19,792.0 &	44.5 &	72.2 &	55.0 &	31.9 &	86.2 \\
\multicolumn{1}{r}{\gold $k$} & 14 &	20,944.0 &	45.7 &	69.1 &	55.0 &	31.6 &	84.3 \\

\bottomrule
\end{tabular}%
}
\caption{PDP and PCFP results for all languages and models, averaged over 4 runs. Metrics are defined in \cref{metrics}. An refers to the Analogy metric and LE to the Lexicon Expansion metric.}
\label{res}
\end{table}

\begin{table*}
    \centering
    \resizebox{0.85\textwidth}{!}{%
    \begin{tabular}{ccccc|ccccc|c} \toprule
    \multicolumn{5}{c}{\cell{sg}} & \multicolumn{5}{c}{\cell{pl}}\\
    \cell{nom} & \cell{gen} & \cell{dat} & \cell{acc} & \cell{abl} &
    \cell{nom} & \cell{gen} & \cell{dat} & \cell{acc} & \cell{abl} & \textbf{Gloss} \\
    \hline
    \word{serv-us} & \word{i} & \word{o} & \word{um} & \word{o} & \word{i} & \word{orum} & \word{is} & \word{os} & \word{is} & ``slave.\cell{m}''\\
    \word{serv-a} & \word{ae} & \word{ae} & \word{am} & \word{a} & \word{ae} & \word{arum} & \word{is} & \word{as} & \word{is} & ``slave.\cell{f}''\\
    \word{frat-er} & \word{ris} & \word{ri} & \word{rem} & \word{re} & \word{res} & \word{rum} & \word{ribus} & \word{res} & \word{ribus} & ``brother'' \\ \bottomrule
    \end{tabular}
    }
    \caption{Suffixal exponents for each cell in the paradigm of three Latin nouns from different inflection classes.}
    \label{sampleLat}
\end{table*}
\cref{res} shows results for two versions of our benchmark system: \bench, as described in \cref{model}, and \gold $k$, with the number of cells oracularly set to the ground truth.
For reference, we also report a supervised benchmark, \superbad, which assumes a gold grid as input, then solves the PCFP exactly as the benchmark does. In terms of the PDP, clustering assigns lexicon forms to paradigms (46--82\%) more accurately than to cells (26--80\%). 
Results are high for English, which has the fewest gold cells, and lower elsewhere.
In German, Latin, and Russian, our benchmark proposes nearly as many cells as \gold $k$, thus performing similarly.
For English, it overestimates the true number and performs worse.
For Arabic, it severely underestimates $k$ but performs better, likely due to the orthography: without diacritics, the three case distinctions become obscured in almost all instances.
In general, fixing the true number of cells can be unhelpful because syncretism and the Zipfian distribution of cells creates situations where certain gold cells are too difficult to detect.
Allowing the system to choose its own number of cells lets it focus on distinctions for which there is sufficient distributional evidence.

As for the PCFP, our benchmark system does well on lexicon expansion and poorly on the analogy task.
While lexicon expansion accuracy (50--86\% compared to 72--97\% for \superbad) shows that the benchmark captures meaningful inflectional trends, analogy accuracy demonstrates vast room for improvement in terms of consistently organizing cell-realizations across paradigms.
English is the only language where analogy accuracy is within half of \superbad's upper bound.
A major reason for low analogy accuracy is that forms, despite being clustered into paradigms well, get assigned to the wrong cell, or the same gold cell gets misaligned across paradigms from different inflection classes.
We discuss this phenomenon in more detail below.


\begin{table}
    \centering
    \resizebox{0.95\columnwidth}{!}{%
    \begin{tabular}{lll} \toprule
    \textbf{Cell} & \textbf{Interpretations} & \textbf{Suffix}\\ \midrule
      0 & \cell{acc.sg} (.51), \cell{gen.pl} (.45) & \word{um}\\
      1   & \cell{acc.pl} (.71), \cell{nom.pl} (.27) & \word{s} \\
      2   & \cell{acc.sg} (.99) & \word{m} \\
      3   & \cell{abl.pl} (.52), \cell{gen.sg} (.40) & \word{is} \\
      4   & \cell{nom.sg} (.39), \cell{abl.sg} (.36) & \word{a} \\
      5   & \cell{abl.sg} (.62), \cell{nom.sg} (.36) & \word{o} \\
      6   & \cell{gen.sg} (.46), \cell{dat.sg} (.30) & \word{i} \\
      7   & \cell{abl.pl} (.77), \cell{dat.pl} (.25) & \word{s} \\
      8   & \cell{nom.sg} (.67), \cell{abl.sg} (.22) & $\varnothing$ \\
      9   & \cell{abl.sg} (.936) & \word{e} \\
      10  & \cell{abl.sg} (.5), \cell{gen.sg} (.28) & \word{e} \\
      11  & \cell{nom.sg} (.87), \cell{acc.pl} (.16) & \word{us}\\  \bottomrule
    \end{tabular}
    }
    \caption{System clustering of Latin nouns.}
    \label{tab:latin_result}
\end{table}

\subsection{Latin Noun Error Analysis}
A detailed analysis of Latin nouns (also analyzed by \newcite{stump2015complexity} and \newcite{beniamine2018inferring}) reveals challenges for our system.
\cref{sampleLat} shows the inflectional paradigms for three Latin nouns exemplifying different inflection classes, which are mentioned throughout the analysis.
In keeping with the UD standard, there are no diacritics for long vowels in the table.

One major challenge for our system is that similar affixes can mark different cells in different inflection classes, e.g. the \textsc{acc.sg} of \textit{servus} ``slave.\textsc{m}'' ends in \textit{um}, as does the \textsc{gen.pl} of \textit{frater} ``brother''.
\cref{tab:latin_result} shows 
system-posited cells, the gold cells they best match to, and the longest suffix shared by 90\% of their members.
The system is often misled by shared affixes, e.g., cell 0 is evenly split between \textsc{acc.sg} and \textsc{gen.pl}, driven by the suffix \textit{um} (cells 3 (\textit{is}) and 4 (\textit{a}) suffer from this as well). 
This kind of confusion could be resolved with better context modeling, as each distinct underlying cell, despite sharing a surface affix, occurs in distinct distributional
contexts.
We observe that the current system does not appear to make use of context to handle some misleading suffixes.
Cell 7 correctly groups \textsc{abl.pl} forms marked with both \textit{is} and \textit{ibus}, excluding other suffixes ending in \textit{s}. 
Similarly, cell 8 contains \textsc{nom.sg} forms with heterogeneous endings, e.g., \textit{r}, \textit{ix} and \textit{ns}.

In some cases, the system misinterprets derivational processes as inflectional, combining gold paradigms.
Derivational relatives \textit{servus} and \textit{serva}, male and female variants of ``slave'', are grouped into one paradigm, as are \textit{philosophos} ``philosopher'' and \textit{philosophia} ``philosophy.''
In other cases, cell clustering errors due to shared suffixes create spurious paradigms.
After falsely clustering gold paradigm mates \word{servum} (\cell{acc.sg}) and \textit{servorum} (\cell{gen.pl}) into the same cell, we must assign each to separate paradigms during paradigm clustering.
This suggests clustering cells and paradigms jointly might avoid error propagation in future work.

We also find that clustering errors lead to PCFP errors.
For \word{servus/a}, the neural reinflector predicts \textit{servibus} in cell 8 with a suffix from the wrong inflection class, yet the slot should not be empty in the first place.
The correct form, \textit{servis}, is attested, but was mistakenly clustered into cell 3.

\subsection{Benchmark Variations Analysis}

\begin{table}
\resizebox{0.99\columnwidth}{!}{%
\setlength{\tabcolsep}{3pt}
\begin{tabular}{rc|ccc|cc}
\toprule
    \multicolumn{1}{c}{\multirow{2}{*}{
    }} & \multicolumn{1}{c}{} & \multicolumn{3}{c}{\textbf{PDP}} & \multicolumn{2}{c}{\textbf{PCFP}} \\
\multicolumn{2}{r}{Paradigms} & $\Fcell$ & $\Fpar$ & $\Fgrid$ & An & LE \\  [1mm]

\multicolumn{7}{c}{\textbf{Arabic nouns} -- 27 cells} \\
\hline
\multicolumn{1}{r}{Gold $k$} & 4,930.3 &	25.9 &	46.4 &	33.1 &	16.1 &	57.2 \\
\cline{1-1}
\multicolumn{1}{r}{larger corpus} & 5,039.5 &	29.1 &	37.5 &	32.8 &	20.4 &	49.2 \\
\multicolumn{1}{r}{smaller corpus} & 5,004.0 &	18.8 &	37.7 &	24.9 &	9.5 & 42.1 \\
\cline{1-1}
\multicolumn{1}{r}{no affix bias} & 4,860.3 &	21.5 &	47.7 &	29.7 &	16.3 &	43.5 \\
\multicolumn{1}{r}{no window bias} & 4,978.5 &	24.0 &	47.5 &	31.8 &	17.6 &	55.8 \\
\cline{1-1}
\multicolumn{1}{r}{$\omega(x,c)=1$} & 3,685.0 & &	34.4 &	28.8 &	5.2 & 35.5 \\
\multicolumn{1}{r}{$\omega(x,c)=0$} & 1,310.5 & &	10.0 &	13.9 &	0.1 &	5.8 \\
\cline{1-1}
\multicolumn{1}{r}{random sources} &  &  &  &  & 16.3 & 55.9 \\
[1mm]

\multicolumn{7}{c}{\textbf{Latin nouns} -- 12 cells} \\
\hline
\multicolumn{1}{r}{Gold $k$} & 3,749.0 &	39.9 &	71.6 &	51.3 &	17.5 &	72.6 \\
\cline{1-1}
\multicolumn{1}{r}{larger corpus} & 3,529.5 &	42.8 &	79.1 &	55.5 &	16.2 &	69.9 \\
\multicolumn{1}{r}{smaller corpus} & 4,381.5 & 30.7 &	49.1 & 37.8 &	14.6 &	51.1 \\
\cline{1-1}
\multicolumn{1}{r}{no affix bias} & 3,906.8 &	37.1 &	68.2 &	48.1 &	22.7 &	66.6 \\
\multicolumn{1}{r}{no window bias} & 3,756.5 &	42.0 &	71.2 &	52.8 &	17.9 &	70.9 \\
\cline{1-1}
\multicolumn{1}{r}{$\omega(x,c)=1$} & 3,262.5 & &	67.1 &	49.6 &	11.0 &	52.9 \\
\multicolumn{1}{r}{$\omega(x,c)=0$} & 1,333.3 & &	26.3 &	31.7 &	0.7 &	7.1 \\
\cline{1-1}
\multicolumn{1}{r}{random sources} &  &  &  &  & 16.5 & 72.3 \\

\bottomrule
\end{tabular}%
}
\caption{Benchmark variations demonstrating the effects of various factors, averaged over 4 runs.}
\label{ablationTable}
\end{table}

\cref{ablationTable} evaluates variants of the benchmark to determine the contribution of several system--task components in Arabic and Latin.
We consider augmenting and shrinking the corpus. 
We also reset the FastText hyperparameters used to achieve a morphosyntactic inductive bias to their default values (no affix/window bias) and consider two constant exponent penalty weights ($\omega(x_f,c)=1$ and $\omega(x_f,c)=0$) instead of our heuristic weight defined in \cref{penalty}. Finally, we consider selecting random sources for PCFP reinflection instead of identifying reliable sources.
For all variants, the number of cells is fixed to the ground truth.

\paragraph{Corpus Size}
We consider either using a smaller corpus containing only the UD subset, or using a larger corpus containing 15 (Latin) or 100 (Arabic) million words from additional supplementary sentences.
As expected, performance decreases for smaller corpora, but it does not always increase for larger ones, potentially due to domain differences between UD and the supplemental sentences.
Interestingly, $\Fcell$ always increases with larger corpora, yet this can lead to worse $\Fpar$ scores, 
more evidence of error propagation that might be avoided with joint cell--paradigm clustering.

\paragraph{Embedding Morphosyntactic Biases}
Targeting affix embeddings by shrinking the default FastText character $n$-gram sizes seems to yield a much more significant effect than shrinking the context window.
In Latin, small context windows can even hurt performance slightly, likely due to extremely flexible word order, where agreement is often realized over non-adjacent words.

\paragraph{Exponent Penalties}
When clustering paradigms with the penalty weight $\omega(x,c)=1$, (which is equivalent to just running the first pass of paradigm clustering), we see a steep decline in performance as opposed to the proposed heuristic weighting.
It is even more detrimental to not penalize exponents at all (i.e., $\omega(x,c)=0$), but maximize the base characters in paradigms without concern for size or likelihoods of exponents. Given allomorphic variation and multiple inflection classes, we ideally want a penalty weight which is lenient to more than just the single most likely exponent, but without supervised data, it is difficult to determine when to stop being lenient and start being harsh in a language agnostic manner.
Our choice to be harsh by default proposes fewer false paradigm mates, yielding less noisy input to train the reinflection model.
In a post-hoc study, we calculated \gold $k$ PCFP scores on \emph{pure} analogies only, where the first three attested forms were assigned correctly during clustering.
Pure analogy PCFP scores were still closer to \gold $k$'s performance than \superbad's for all languages.
This suggests most of the gap between \gold $k$ and \superbad is due to noisy training on bad clustering assignments, not impossible test instances created by bad clustering assignments.
This supports our choice of harsh penalties and suggests future work might reconsider clustering decisions given the reinflection model's confidence.

\paragraph{Reinflection Source Selection}
During reinflection, feeding the Transformer random sources instead of learning the most reliable source cell for each target cell slightly hurts performance. The margin is small, though, as most paradigms have only one attested form.
In preliminary experiments, we also tried jointly encoding all available sources instead of just the most reliable, but this drastically lowers performance.

 \label{results}

\section{Conclusion}

We present a framework for the paradigm discovery problem, in which words attested in an unannotated corpus are analyzed according to the morphosyntactic property set they realize and the paradigm to which they belong.
Additionally, unseen inflectional variants of seen forms are to be predicted.
We discuss the data required to undertake this task, a benchmark for solving it, and multiple evaluation metrics.
We believe our benchmark system represents a reasonable approach to solving the problem based on past work and highlights many directions for improvement, e.g. joint modeling and making better use of distributional semantic information. 

\section*{Acknowledgments}
The authors would like to thank the members of New York University Abu Dhabi's CAMeL Lab, Marie-Catherine de Marneffe, Eleanor Chodroff, Katharina Kann, and Markus Dreyer.
We acknowledge the support of the High Performance Computing Center at New York University Abu Dhabi.
Finally, we wish to thank the anonymous reviewers at EMNLP 2019 and ACL 2020 for their feedback.

\bibliography{acl2020}
\bibliographystyle{acl_natbib}

\end{document}